# A Novel Feature Extraction Method for Scene Recognition Based on Centered Convolutional Restricted Boltzmann Machines

Jingyu Gao[1], Jinfu Yang[1]*, Guanghui Wang[2] and Mingai Li[1]

*1 Department of Control & Engineering Beijing University of Technology, NO.100 Chaoyang district, Beijing, 100124, P.R.China.*

*2 Department of Electrical Engineering & Computer Science University of Kansas, Lawrence, KS 66045-7608, USA*

## ABSTRACT

Scene recognition is an important research topic in computer vision, while feature extraction is a key step of object recognition. Although classical Restricted Boltzmann machines (RBM) can efficiently represent complicated data, it is hard to handle large images due to its complexity in computation. In this paper, a novel feature extraction method, named Centered Convolutional Restricted Boltzmann Machines (CCRBM), is proposed for scene recognition. The proposed model is an improved Convolutional Restricted Boltzmann Machines (CRBM) by introducing centered factors in its learning strategy to reduce the source of instabilities. First, the visible units of the network are redefined using centered factors. Then, the hidden units are learned with a modified energy function by utilizing a distribution function, and the visible units are reconstructed using the learned hidden units. In order to achieve better generative ability, the Centered Convolutional Deep Belief Networks (CCDBN) is trained in a greedy layer-wise way. Finally, a softmax regression is incorporated for scene recognition. Extensive experimental evaluations using natural scenes, MIT-indoor scenes, and Caltech 101 datasets show that the proposed approach performs better than other counterparts in terms of stability, generalization, and discrimination. The CCDBN model is more suitable for natural scene image recognition by virtue of convolutional property.

## 1. Introduction

Scene recognition is an important task for many practical applications, such as robot navigation, location, and map construction. Generally speaking, scene recognition consists of two basic procedures: feature extraction and classification. Feature extraction is a key step to object recognition. In recent years, a number of feature extraction approaches for scene recognition have been proposed. Sande[1] introduced color descriptors to increase illumination invariance and discriminative power in a structured way. They employed taxonomy to analytically show the invariant properties of color descriptors and utilized two benchmarks from an image domain and a video domain, respectively, to assess the distinctiveness of color descriptors. Quatton and Torralba [2] proposed a prototype model that defined a mapping between object images and scene labels, which could capture similar objects in related scenes. Lu[3] presented a two-level algorithm by firstly distinguishing the scenes into outdoor and indoor, and then classifying the scenes into multiple categories. Brown and Susstrunk [4] proposed a multi-spectral scale-invariant feature transform (SIFT) approach for scene recognition which combined a kernel-based classifier with SIFT[5]. In addition to the above feature extraction approaches, Histogram of Oriented Gradient (HOG) [6], Gradient Location-Orientation Histogram (GLOH) [7] and so on, received more attention than the global features in recent years. However, local features have two limitations in practice. First, they more or less ignore some hidden characteristics of images, as a consequent, not all internal information is included in the features. Second, the extraction procedure with humans' intervention could not sufficiently express all details in the images. In contrast to the local features, the global features extracted by deep learning, a unsupervised learning algorithms, are more comprehensive and recognizable[8].

* Corresponding author. Tel.: +86-10-67396309; e-mail: jfyang@bjut.edu.cn

In deep networks, detector units are embedded in every layer, and the higher layers, by detecting more complicated features, receive the simple features detected by lower layers. In 2006, Hinton and Salakhutdinov [8] proposed a deep learning model – Deep Belief Network (DBN) and the greedy layer-wise training approach. The deep learning models, such as DBN [9], Deep Boltzmann machines (DBM) [10], and Convolutional Neural Network (CNN) [11] have mushroomed in recent years, and have been widely applied in robot navigation [12], face recognition [13], and document modeling [14]. Compared to previous artificial neural networks, deep learning based approaches avoid the over-fitting problem; and the residual error is reduced by using a layer to layer transfer. Restricted Boltzmann machines (RBM) is the basis of deep learning models such as the DBN and the DBM that could be used in handwriting digit recognition, object recognition [15], and human motion capture [16]. Thus, more and more researchers paid attention to the variant of the Restricted Boltzmann machines (RBM). For example, Ranzato [17] proposed a factored 3-way RBM to control the covariance of visible layers using the factors between neighboring layers. Srivastava and Salakhutdinov [14] presented an over-replicated softmax model to increase the number of hidden layers without adding more parameters through coping with the weights of the lower layers.

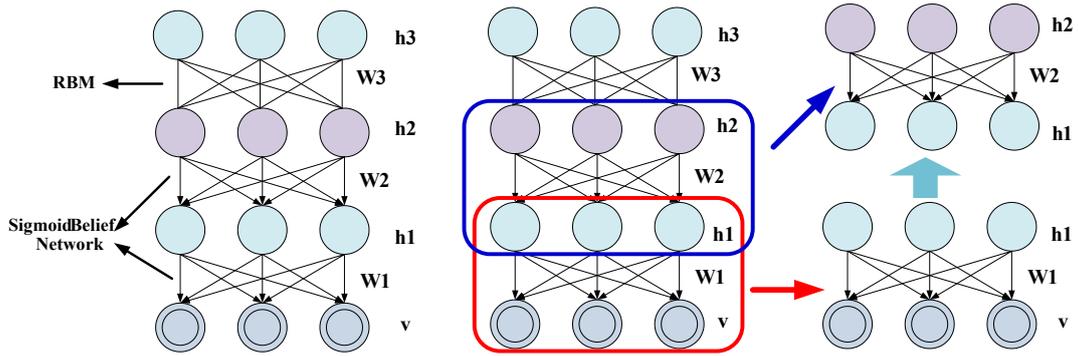

**Fig.1.** Left: the architecture of DBN. The lower two layers are sigmoid belief networks, and the others consist of RBM as described in Section 2.1. Right: the learning procedure of DBN. The network is pre-trained layer by layer, and the output of the last layer is regarded as the initialization of next layer.

Although the RBM and its variants can extract high quality features, scaling them to the size of natural images, such as 200x200 pixels, they are difficult to extract accurate features due to large numbers of visible and hidden units. The Convolutional Restricted Boltzmann Machines (CRBM), which can generate pint-sized two-dimensional weights through convolution, is a more efficient generative model for full-sized natural images. It is a translation-invariant hierarchical generative model which supports both top-down and bottom-up probabilistic inference[18]. The CRBM and the Convolutional Deep Belief Networks (CDBN), which is stacked by CRBMs, have been successfully applied in handwritten digits recognition, object classification [18], and pedestrian detection [19], however, they ignored the sources of instability during the learning procedure. Centered factors were introduced in DBM to reduce the instability caused by approximation and replacement [20].

In this paper, we propose a novel feature extraction algorithm for scene recognition named Centered Convolutional Restricted Boltzmann Machines (CCRBM) based on CRBM by integrating the centered factors into the learning procedure of the CRBM model. First, the visible units are redefined using the input data and the centered factors. Then, the hidden units are learned with a modified energy function by utilizing a distribution function; and the visible units are reconstructed using the learned hidden units. Next, in order to get a better generative ability, Centered Convolutional Deep Belief Networks (CCDBN) is trained in a greedy layer-wise way. Finally, the softmax regression, which is a variant of the logistic regression, is incorporated to perform scene recognition tasks. The main contributions of this paper are as follows:

(1) Compared to the standard CRBM, the CCRBM effectively reduces the sources of the instability caused by the model structure and the approximation in the learning procedure.

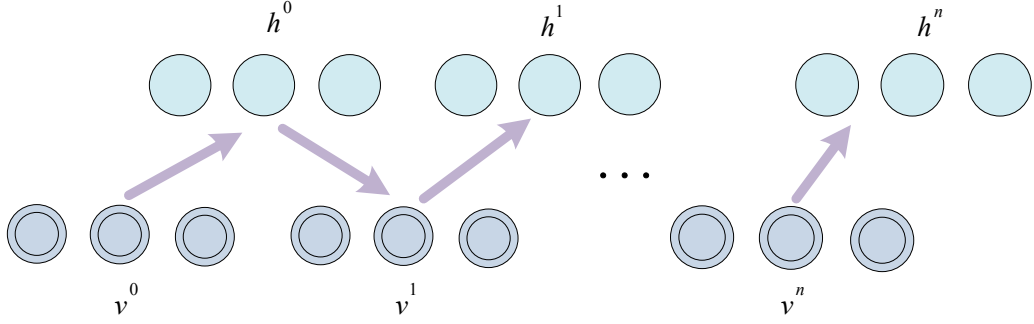

**Fig.2.** The procedure of Gibbs sampling. Given $v^0$, the first step is computing $h^0$. Next, compute $v^1$ from equation 5 to complete one step Gibbs sampling. $h^i$ and $v^i$ are calculated in the same way until $i=n$.

(2) Different to the DBN and DBM, which can only handle small images like handwritten digit images, the proposed CCDBN can train large images, such as natural scene images. Thus, the CCDBN is more suitable for scene recognition.

The remainder of this paper is structured as follows. Section 2 introduces the structure and learning procedure of the RBM and DBN. Section 3 presents the proposed CRBM with the centered factors. The centered factors consist of redefining the energy function of the CRBM, which improves the condition of the optimization problem and promotes the emergence of complicated structures in the CDBN. The softmax classifier, which can be used in multi-class problem, is discussed in Section 4. The performance of the CCRBM with the softmax classifier is evaluated and discussed in Section 5 using fifteen scene categories. Finally, the paper is concluded in Section 6.

## 2. Prior Works

In this section, we briefly present some background on the RBM and DBN to facilitate the understanding of this paper.

### 2.1 Restricted Boltzmann Machines

Restricted Boltzmann Machines (RBM) [15] is a bipartite graph with two layers. It consists of visible units $v \in \{0,1\}^D$ and hidden units $h \in \{0,1\}^P$, where every visible unit is connected to all hidden units by a weight matrix $w$, as shown in Fig.1, while the units do not connect with each other within the same layer.

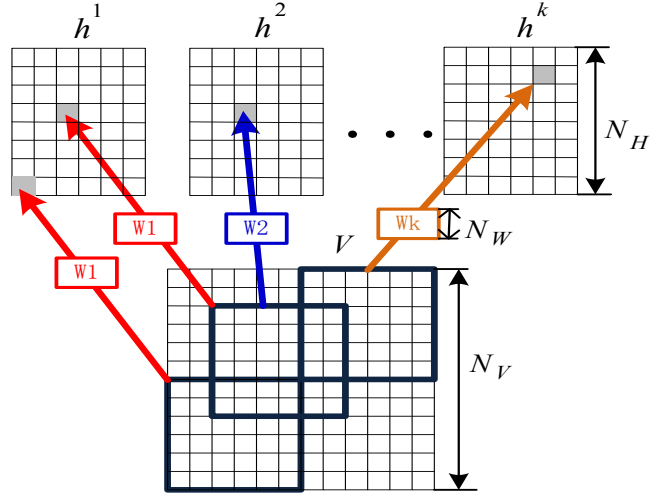

**Fig.3.** A CRBM with $k$ groups of $3\times 3$ filters. The feature maps $h^1$ to $h^k$ by $w$ ergodic convolution on visible layer, instead of visible variable matrix multiplying hidden variable matrix in the standard RBM.

The probability of the visible variables in an RBM with the parameter set $\theta$ in accordance with a joint energy $E(v,h;\theta)$ is defined as:

$$P(v,h) = \frac{1}{Z(\theta)} \sum_h \exp(-E(v,h,\theta)) \quad (1)$$

where $v$ and $h$ denote the vectors of the visible and hidden variables, respectively; $exp(x)$ represents the power of a constant $e$; and $Z(\theta)$ is a normalization constant defined by

$$Z(\theta) = \sum_v \sum_h \exp(-E(v,h,\theta)) \quad (2)$$

where $E(v,h,\theta)$ is a free energy function of the RBM with

$$E(v,h;\theta) = -\sum_{i,j} v_i w_{i,j} h_j - \sum_i b_i v_i - \sum_j c_j h_j \quad (3)$$

where $i$ and $j$ are the sequence numbers of the visible and hidden units; the variable $\theta$ is a set of model parameters $\theta = \{w, b_i, c_j\}$, where $b_i$ and $c_j$ denote the hidden and visible unit biases, respectively; and $w$ denotes the connection between the pairs of visible and hidden units.

Inferring the distribution of these hidden variables is easy since there is no connection between the hidden variables.

$$P(h_j|v) = \sigma\left(c_j + \sum_i v_i w_{ij}\right) \tag{4}$$

$$P(v_i|h) = \sigma\left(b_i + \sum_j w_{ij} h_j\right) \tag{5}$$

where $\sigma(x) = 1/(1+e^{-x})$ is a sigmoid function.

## 2.2 Learning Procedure of Restricted Boltzmann Machines

The parameters of the RBM can be learned by maximizing the likelihood. The derivative of the log-likelihood is

$$\frac{\partial}{\partial \theta} L(\theta) = -\left\langle \frac{\partial E(\mathrm{v};\theta)}{\partial \theta} \right\rangle_{data} + \left\langle \frac{\partial E(\mathrm{v};\theta)}{\partial \theta} \right\rangle_{model} \tag{6}$$

where $E(\mathrm{v};\theta)$ is given in equation (3); and $\langle \cdot \rangle_{data}$ and $\langle \cdot \rangle_{model}$ denote the expected values of the visible vector $v$ in regard to the data and the model distribution, respectively. Unfortunately, it is difficult to compute the expected values since it involves an exponential number of terms. Hinton [21] proposed an alternative objective function, named Contrastive Divergence (CD), to solve the problem by maximizing the likelihood during learning.

The update rule for parameter $w_{ij}$ in the CD function is defined as:

$$w_{ij} = w_{ij} + \eta\left(\langle v_i^0 h_j^0 \rangle - \langle v_i^n h_j^n \rangle\right) \tag{7}$$

where $\eta$ is a learning rate; $v^0$ is estimated from the observed data distribution; $h^0$ is acquired according to equation (4); $v^n$ is recovered from the sampled data by $n$ steps of Gibbs sampling; $h^n$ is recovered from equation (4) based on $v^n$; and $\langle \cdot \rangle$ denotes the expected value. The values of $v^n$ and $h^n$ are computed using alternating Gibbs sampling, as shown in Fig.2.

During the learning process of the CD algorithm, Gibbs sampling [9] is initialized by the input data, and the algorithm runs a few steps to obtain an approximation of the model distribution. The first step is to compute the $h^0$ with a given $v^0$ from equation (4), while the value of $P(h_0|v_0)$ is regarded the same as that of the $h^0$. Then, the $v^1$ is computed according to equation (5). Based on the above process of Gibbs sampling, $h^i$ and $v^i$ can be calculated until $i=n$, where the variable $i$ stands for the index of different moment rather than the number of layers.

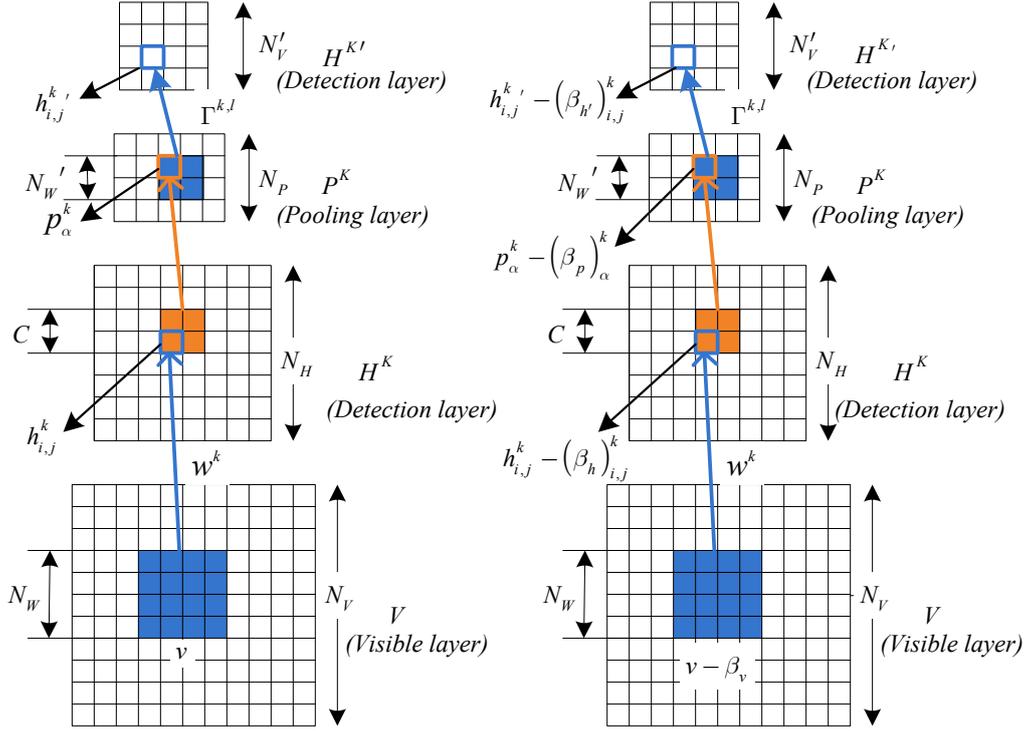

**Fig.4.** CDBN and CCDBN with probabilistic max-polling. It only displays the group $k$ of detection layer $P^k$ and pooling layer $H^k$, $H^{k'}$. The lower three layers are CRBM (left) and CCRBM (right) with the visible layer, detection layer and pooling layer, respectively. Left image shows the structure of CDBN and the right one is the structure of CCDBN. See Section 3 for details.

## 2.3 Deep Belief Networks

Deep Belief Networks (DBN) is a hierarchical generative model and is stacked by sigmoid belief networks and RBMs (as shown in Fig.1). There are connections between adjacent two layers, while the units in the same layer are not connected with each other [15]. The parameters of the bottom layer $\theta_1$ is learned by training an RBM between $v$ and $h_1$ with the steps described in Section 2.2. After learning the first layer, the parameter $\theta_1$ is fixed, while the probability $P(h_1|v;\theta_1)$ of the hidden units in the first layer is taken as the input of the second layer $h_1$ to train the second RBM between $h_1$ and $h_2$. Similarly, other layers can be added to the model through the same procedure. The above procedure of training the DBN model is illustrated in Fig.1.

## 3. Learning Centered Convolutional Restricted Boltzmann Machines

In this section, we will introduce a novel approach, named CCRBM, by combining centered factors into the learning process in order to reduce the source of instabilities from approximation and structure.

### 3.1 Convolutional Restricted Boltzmann Machines

From the above analysis, it is easy to find that the spatial relationship between different image patches is not considered in the standard RBM since the input image is treated as a vector. As a result, the features extracted from adjacent patches become independent [9]. On the other hand, the standard RBM cannot be applied to large images. For example, if the input image is $200 \times 200$ pixels, the visible variable will be a 40000-vector. It is hard for the RBM to handle such a large input due to computational complexity.

To solve these problems, an extension of RBM, called Convolutional Restricted Boltzmann machines (CRBM), was introduced in [18]. The weight matrix of the CRBM between the visible and hidden layers for the local area is shared on entire image to generate spatial structure of adjacent patches. The CRBM includes two layers as the RBM: a visible layer $v$ and a hidden layer $h$, as shown in Fig.3. The visible layer consists of an $N_v \times N_v$ array of binary units and the hidden layer consists of $K$ "groups", where each group is an $N_H \times N_H$ array of binary units. Thus, there are $N_H^2 K$ hidden units altogether [19]. Each group $N_W \times N_W$ is connected with an filter, where $N_W = N_V - N_H + 1$. The filter weight is shared by all the hidden units in this group, while each hidden group has bias $b_k$ and $c$, which denote the hidden layer bias and the visible layer, respectively.

In the CRBM, the energy function $E(v,h)$ is defined as:

$$E(v,h) = -\sum_{k=1}^{K} \sum_{i,j=1}^{N_H} \sum_{s=1}^{N_W} h_{ij}^k w_{rs}^k v_{i+r-1,j+s-1} - \sum_{k}^{K} b_k \sum_{i,j=1}^{N_H} h_{ij}^k - c \sum_{i,j=1}^{N_V} v_{ij} \tag{8}$$

So energy function is redefined as:

$$P(v,h) = \frac{1}{Z(\theta)} \sum_{h} \exp(-E(v,h;\theta)) \tag{9}$$

$$E(v,h) = -\sum_{k=1}^{K} h^k \bullet (w^k * v) - \sum_{k=1}^{K} b_k \sum_{i,j} h_{i,j}^k - c \sum_{i,j} v_{ij} \tag{10}$$

where '*' denotes convolution, and '•' represents the trace, for example $A \bullet B = trA^T B$. Just like the standard RBM, the conditional distribution of the CRBM is defined by block Gibbs sampling:

$$P\left(h_{ij}^k = 1 | v\right) = \sigma\left(\left(w^k * v\right)_{ij} + b_k\right) \qquad (11)$$

$$P\left(v_{ij} = 1 | h\right) = \sigma\left(\left(\sum_k w^k * h^k\right)_{ij} + c\right) \qquad (12)$$

where $h_{ij}^k$ represents the *i-th* row and *j-th* unit of the *k-th* group in the hidden layer. $v_{ij}$ represents the *i-th* row and *j-th* unit in the visible layer. $w^k$ is the *k-th* group weight.

## 3.2 Training Procedure of Centered Convolutional Restricted Boltzmann Machines

### 3.2.1 Training Procedure of CRBM in General Way

After understanding the structure of the CRBM, the learning procedure will be introduced in this section. Like the RBM, the real power of the CRBM emerges when it is stacked to form a Convolutional Deep Belief Network (CDBN) [19]. A probabilistic max-pooling operation is used to change the architecture of CRBM to learn high-level representation. As shown in Fig.4, there are two layers in the original hidden layers: detection layer, whose results are calculated by convolving a feature detector of the previous layer, and the pooling layer which shrinks the results of the detection layer by a constant factor. Each unit in the pooling layer is the max probability of the units in a small area (such as $4 \times 4$ pixels or $10 \times 10$ pixels) of the detection layer. Shrinking the activation with max-pooling enables higher-layer representations to be constant to small translations of the observed data and reduce the computational burden. The energy function of this simplified probabilistic max-pooling CRBM is defined as follows:

$$E(v, h) = -\sum_k \sum_{i,j} \left( h_{i,j}^k \left( w^k * v \right)_{i,j} + b_k h_{i,j}^k \right) - c \sum_{i,j} v_{i,j}$$
$$subj.to \quad \sum_{(i,j) \in B_\alpha} h_{i,j}^k \leq 1, \forall k, \alpha \qquad (13)$$

where *k-th* group receives the bottom-up signal from layer $v$ as follows:

$$I\left(h_{i,j}^k\right) \triangleq b_k + \left(w^k * v\right)_{ij} \qquad (14)$$

Suppose $h_{i,j}^k$ is a hidden unit included in the block, i.e., $(i,j) \in B_\alpha$, the increase of energy is $-I(h_{i,j}^k)$, caused by the units $h_{i,j}^k$, then, the conditional probability is

$$P(h_{i,j}^k = 1|v) = \frac{\exp(I(h_{i,j}^k))}{1 + \sum_{(i',j') \in B_\alpha} \exp(I(h_{i',j'}^k))} \qquad (15)$$

$$P(p_\alpha^k = 0|v) = \frac{1}{1 + \sum_{(i',j') \in B_\alpha} \exp(I(h_{i',j'}^k))} \qquad (16)$$

Given the hidden layer $h$, the visible layer $v$ can be sampled in the same way as described in Section 3.1.

### 3.2.2 Centered Convolutional Restricted Boltzmann Machines

In the above learning procedure, there are two sources of instability [20]: (1) Approximation instability: the noisy gradient is resulted from approximation sampling procedure, causing deviation from the true value. For example, the equation (7) instead of the equation (6), and the $P(h|v)$ and $P(v|h)$ respectively represent the values of $h$ and $v$. (2) Structural instability: as identified in [22], the weight matrix $w$ in the first step of the Boltzmann machines is a global bias instead of dependencies between each units as expected. This is prominently problematic for the Boltzmann machines with several layers such as the DBN and the CDBN. The bias formed in the hidden units forming a bias can speed up learning at the beginning but it will eventually destroy the learning between pairs of hidden units.

The centered factors are introduced to relieve these sources of instability by guaranteeing that unit activations are centered by intervention the computation of the gradient. Centering avoids using a global bias, which reduces the noise of the learning procedure.

Centered factors can be acquired by reformulating the energy function, and the CCRBM of the energy function is redefined as:

$$E(v,h) = -\sum_k \sum_{i,j} \left( h_{i,j}^k \left( w^k * (v - \beta_v) \right)_{i,j} + b_k \left( h_{i,j}^k - (\beta_h)_{i,j}^k \right) \right) - c \sum_{i,j} \left( v_{i,j} - (\beta_v)_{i,j} \right) \qquad (17)$$

where the new variables $\beta_v$ and $\beta_h$ stand for the offset of the visible units and the hidden units, respectively. Setting $\beta_{v_0} = \sigma(b_0)$ and $\beta_{h_0} = \sigma(c_0)$, where $b_0$ and $c_0$ are the initial hidden and visible layer biases, ensuring that units are initially centered. From the equation (17), we can rewrite the conditional probability of the CRBM as

$$P(p_\alpha^k = 0|v) = \frac{1}{1+\sum_{(i',j')\in B_\alpha} \exp(\Delta I(h_{i',j'}^k))} \tag{18}$$

$$P(h_{i,j}^k = 1|v) = \frac{\exp(\Delta I(h_{i,j}^k))}{1+\sum_{(i',j')\in B_\alpha} \exp(\Delta I(h_{i',j'}^k))} \tag{19}$$

$$\Delta I(h_{i,j}^k) \triangleq b_k + \left(w^k * (v - \beta_v)\right)_{ij} \tag{20}$$

and the conditional probability of sampling visible units given hidden units is defined by

$$P(v_{ij} = 1|h) = \sigma\left(\left(\sum_k w^k * (h^k - (\beta_h)^k)\right)_{ij} + c\right) \tag{21}$$

The update equations are changed by the new centered factors constraints as follows:

$$\begin{aligned} b' &= b + w * (\langle v - \beta_v \rangle) \\ c' &= c + w * (\langle h - \beta_h \rangle) \\ \beta_h' &= \langle h \rangle, \beta_v' = \langle v \rangle \end{aligned} \tag{22}$$

### 3.3 Hierarchical Probabilistic Inference Based on Centered Factors

Like the CDBN, the CCDBN is defined as the stacked CCRBM and each CCRBM is trained through the way in Section 3.2.2. Then, the greedy layer-wise algorithm described in Section 2.3 is used to train the CCDBN. When a given layer is trained, its weight is frozen, and its activations are used as input for the next layer.

As shown in Fig.4, the architecture of model consists of visible layer $V$, detection layer $H$, pooling layer $P$, and the higher detection layer $H'$. We suppose $H'$ has $K'$ groups of nodes, and there is a shared weight set $\Gamma = \{\Gamma^{1,1}, ..., \Gamma^{K,K'}\}$, where $\Gamma^{k,l}$ is a weight matrix that connects pooling unit $P^k$ to detect $H'^l$ as described in [23].

The energy function for the network with the centered factors has two kinds of potentials: unary terms which are in detection layers for each group, and interaction terms between each pair of neighbor layers.

$$E(v,h,p,h') = -\sum_{k} v \bullet \left( w^k * \left( h^k - (\beta_h)^k \right) \right) - \sum_{k} b_k \sum_{ij} \left( h_{ij}^k - (\beta_h)_{ij}^k \right)$$
$$- \sum_{k,l} p^k \bullet \left( \Gamma^{kl} * \left( h'^l - (\beta_{h'})^l \right) \right) - \sum_{l} b_l' \sum_{ij} \left( h_{ij}'^l - (\beta_{h'})_{ij}^l \right) \quad (23)$$

where $\beta_{h'}$ is the centered factor of the higher detection layer.

The detection layer receives the bottom-up signal from visible layer the same as equation (19), and the pooling layer receives the top-down signal from the new detection layer as follows:

$$\Delta I(p_\alpha^k) \triangleq \sum_{l} \left( \Gamma^{kl} * \left( h'^l - (\beta_{h'})^l \right) \right)_\alpha \quad (24)$$

$$P(h_{i,j}^k = 1 | v, h') = \frac{\exp\left( \Delta I(h_{i,j}^k) + \Delta I(p_\alpha^k) \right)}{1 + \sum_{(i',j') \in B\alpha} \exp\left( \Delta I(h_{i',j'}^k) + \Delta I(p_\alpha^k) \right)} \quad (25)$$

where $(i,j) \in B_\alpha$. The conditional probability with the centered factors is given by

$$P(p_\alpha^k = 0 | v, h') = \frac{1}{1 + \sum_{(i',j') \in B\alpha} \exp\left( \Delta I(h_{i',j'}^k) + \Delta I(p_\alpha^k) \right)} \quad (26)$$

and the conditional probability of sampling visible units given hidden units with the centered factors is the same as equation (21).

**4 Softmax Regression Model**

In order to perform scene recognition, softmax regression was used to classify the scene images in our experiments. Softmax regression is a variant of the logistic regression model which is only applied to binary classification. In the softmax regression setting, we are interested in multi-class classification, so the label $y$ can take on $k$ different values rather than only two. Thus, we have the training set $\{(x^{(1)}, y^{(1)}), ..., (x^{(m)}, y^{(m)})\}$ with $m$ labeled examples and $y^{(i)} \in \{1, 2, 3, ..., k\}$. To estimate the probability of the class labels obtained from $k$ different possible values given a test input $x$, the hypothesis is estimated by the probability $P(y = j | x)$ for each value of $j = 1, 2, ..., k$. Therefore, the hypothesis will yield a $k$-dimensional vector (whose elements sum up to 1) giving $k$ estimated probabilities. Specifically, the hypothesis function $h_{\theta(x)}$ takes the form

$$h_{\theta(x)} = \begin{bmatrix} p(y^{(i)}=1|x^{(i)};\theta) \\ p(y^{(i)}=2|x^{(i)};\theta) \\ p(y^{(i)}=3|x^{(i)};\theta) \\ \vdots \\ p(y^{(i)}=k|x^{(i)};\theta) \end{bmatrix} = \frac{1}{\sum_{j=1}^{k}\exp(\theta_j^T x(i))} \begin{bmatrix} \exp(\theta_1^T x(i)) \\ \exp(\theta_2^T x(i)) \\ \exp(\theta_3^T x(i)) \\ \vdots \\ \exp(\theta_k^T x(i)) \end{bmatrix} \quad (27)$$

where θ is the parameter of the softmax regression model. Notice that the term $\sum_{j=1}^{k}\exp(\theta_j^T x(i))$ normalizes the distribution, so that it sums up to one.

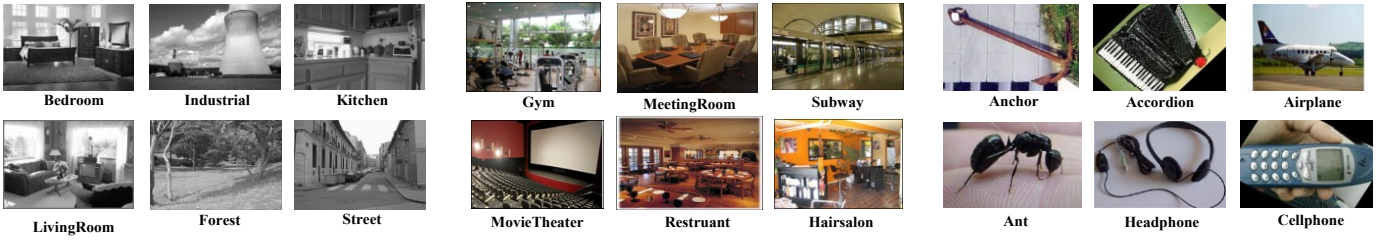

**Fig.5.** Some examples of the fifteen scene categories, MIT-indoor dataset and Caltech 101. The left group is six types of images sampled from the fifteen categories including indoor and outdoor. The middle group is the examples of MIT-indoor scene dataset. The right one is six types of images sampled from Caltech 101

The parameter θ was learnt to minimize the cost function. In equation (28), $1\{\cdot\}$ is an indicator function, i.e., $1\{a\ true\ statement\}=1$, and $1\{a\ false\ statement\}=0$. Thus, the cost function is written as

$$J(\theta) = -\frac{1}{m}\left[\sum_{i=1}^{m}(1-y^{(i)})\log(1-h_\theta(x^{(i)})) + y^{(i)}\log h_\theta(x^{(i)})\right]$$
$$= -\frac{1}{m}\left[\sum_{i=1}^{m}\sum_{j=0}^{1} 1\{y^{(i)}=j\}\log p(y^{(i)}=1|x^{(i)};\ \theta)\right] \quad (28)$$

where θ is random number of initialization $p(y^{(i)}=k|x^{(i)};\theta)$; and

$$P(y^{(i)}=j|x^{(i)};\theta) = \frac{\exp(\theta_j^T x(i))}{\sum_{j=1}^{k}\exp(\theta_j^T x(i))} \quad (29)$$

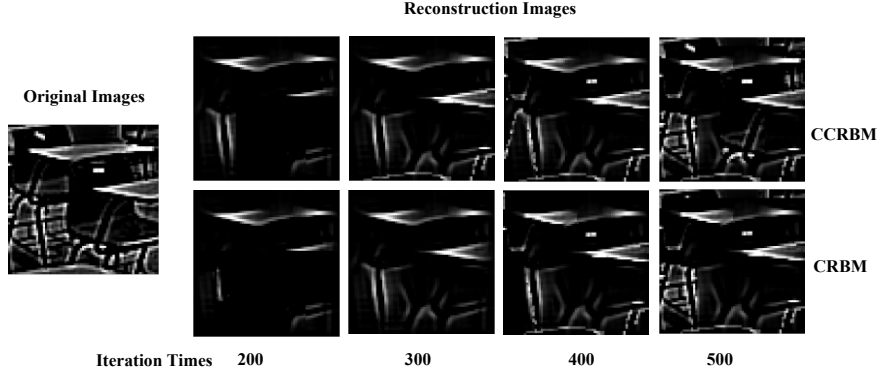

**Fig.6.** The input images and its reconstructed data at iteration 200,300,400, and 500.The left image is original image patches and the right images are the results of reconstruction using the extracted features. The images of top row are results from the CCRBM, and bottom ones are results using the standard CRBM. Obviously, the CCRBM extracts much more information and details.

There is no closed-form solution for the minimization of $J(\theta)$. We adopt an iterative optimization algorithm such as gradient descent or L-BFGS (Limited-memory BFGS) to solve the problem. Taking derivative of equation (28) yields

$$\nabla_{\theta_j} J(\theta) = -\frac{1}{m} \sum_{i=1}^{m} \left[ x^{(i)} \left( 1\{y^{(i)} = j\} - p\left(y^{(i)} = 1 \middle| x^{(i)}; \theta\right)\right)\right] \quad (30)$$

In particular, $\nabla_{\theta_j}$ is a vector, thus, its *l-th* element is the partial derivative of $J(\theta)$ with respect to the *l-th* element of $\theta_j$. During the implementation of gradient descent, $J(\theta)$ we use the update equation $\theta_j = \theta_j - \alpha \nabla_{\theta_j} J(\theta)$ where $j = 1,...,k$. When implementing softmax regression, we typically use a modified version of the cost function as described above. Specifically, it incorporates weight decay. The cost function is redefined by adding the decay term which penalizes large values of the parameters. The new cost function is given by

$$J(\theta) = -\frac{1}{m}\left[\sum_{i=1}^{m}\sum_{j=0}^{1} 1\{y^{(i)} = j\} \log p\left(y^{(i)} = 1 \middle| x^{(i)}; \theta\right)\right] + \frac{\lambda}{2}\sum_{i=1}^{k}\sum_{j=0}^{n} \theta_{ij}^2 \quad (31)$$

The proposed method is summarized as follows:

(1) Define the parameters of the CCDBN, such as the number of layers, the number of units in each layer, iteration times, and initialize the weight matrix between neighboring layers;

(2) Redefine the visible units $v$ by combining centered factors;

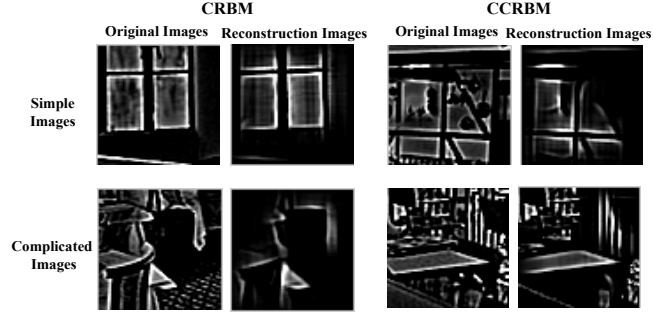

**Fig.7.** The results on indoor scene dataset. The left group is the results of simple input data; and the right group is the results of complicated data.

(3) Train the CCRBM using the following steps:

  a) Given the visible units $v$, calculate $p(h|v)$ and $p(p|v)$ respectively, where $h$ is replaced by the value of $p(h|v)$ and $p$ is replaced by the value of $p(p|v)$;

  b) Compute $p(v|h)$, and setting the reconstructed value $v$ equals to $p(v|h)$;

  c) Update the weight matrix $w$ and the center factors respectively;

  d) Repeat the above steps iteratively until convergence or the maximum number of iterations.

(4) Taking the pooling layer value $p$ of current CCRBM as input of next layer. Repeat step (3) to train other CCRBMs;

(5) When two neighboring CCRBMs have been trained, compute $p(h|v,h')$ and $p(p|v,h')$ and replace $h$ and $p$ with $p(h|v,h')$ and $p(p|v,h')$ respectively, and update $W$;

(6) Utilize the softmax classifier to classify extracted features.

## 5. Experimental Evaluations

Our experiments were performed on a Sony PC with an Intel Core i3 CPU 350 @2.27GHz, and a 6GB random access memory. We report our results on three different datasets: fifteen scene categories dataset[24], MIT-indoor scene dataset [2] and Caltech 101[25].

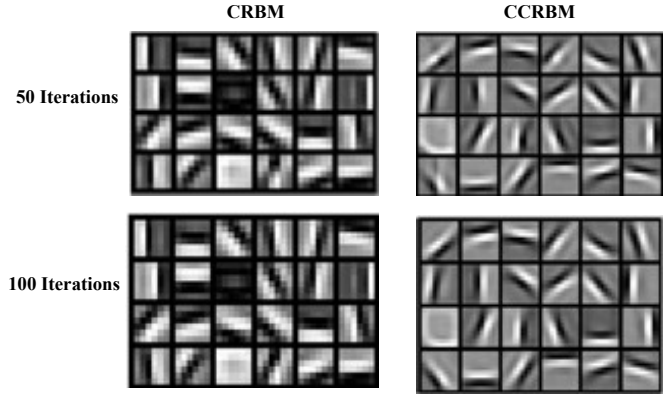

**Fig.8.** The visualization of trained CRBM and CCRBM. Left is the visualization of a trained CRBM with 50 (top) and 100 (bottom) iterations. Right is the visualization of a trained CCRBM with 50 (top) and 100 (bottom) iterations. In the same iteration times, the features extracted by centered CRBM are more distinct and cost less time.

The fifteen class scene dataset is a dataset of fifteen natural scene categories that expands on the thirteen category dataset released in [24]. The two new categories are industrial and store added by Oliva [26]. The sizes of images in the dataset are about 250x300 pixels, with 210 to 410 images per class. This dataset contains a wide range of outdoor and indoor scene environments. The MIT-indoor scene database [2] contains 15620 images of 67 indoor categories. The number of images varies across categories, but there are at least 100 images per category. In Caltech-101 dataset, pictures of objects belong to 101 categories with 40 to 800 images per category. Most categories have about 50 images, collected in September 2003 by Li Fei-Fei [25]. The size of each image is roughly 200x300 pixels. Some examples for every dataset are shown in Fig.5. The images in the left group are the examples of fifteen scene categories dataset; the middle group is examples of MIT-indoor scene dataset and the right one is the example images of Caltech 101.

## 5.1 Experimental Results

We tested the proposed method from the definition of reconstruction of images and visualization of weights using fifteen scene categories dataset, MIT-indoor scene dataset, and Caltech 101 dataset respectively. In the first experiment, we randomly chose 50 images per class from fifteen scene category dataset to assess reconstruction error and convergence rate. In our experiments, the structure of the proposed CCRBM consists of 24 groups of 10x10 pixels filters, and the pooling region is 2x2 pixels. We sampled 100x100 pixels as the input data of the model, and got hidden layer output with 90x90 pixels by averaging the 24 feature maps. The mean values of the pooling layer are regarded as the extracted features of the CCRBM. We conducted experiments using CRBM and CCRBM for iteration times at 200, 300, 400 and 500. The input data of sampled 90x90 image patches and the reconstructed images are shown in Fig.6, where the left image is the original image; and the right group is the reconstructed images. The first row of the reconstruction images is the results of the CCRBM and the second one is the results of the CRBM. As we know, the features extracted by the first layer of multilayer models are always the edges. Obviously, we can discover that the features extracted by the CCRBM are more distinct with more details. For example, from the results of CCRBM, we can identify the back and legs of the chair at 500 iterations and also can easily identify the edges of the table and the chair since 200 iterations. In contrast, one cannot distinguish the parts of the chair under the table even at 500 iterations.

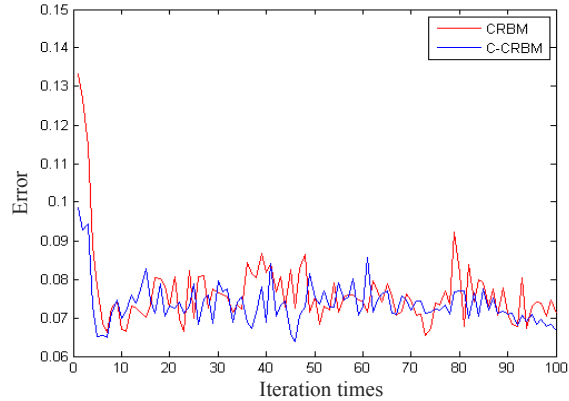

**Fig.9.** The errors of CRBM and CCRBM. The error is defined as the absolute value of the deference between the input and reconstruction data.

We also utilized the same structure of CCRBM to extract features from MIT-indoor scene dataset. We chose 10 categories which are not included in fifteen class scene dataset, such as restaurant, gym, and airport. We randomly chose 40 images per class from the dataset as input data. The training procedure was the same as that in the experiments on fifteen scene categories dataset. The input data and reconstruction results are shown in Fig.7. It is evident that both CCRBM and CRBM can correctly extract features for simple images. For complicated images, however, the CCRBM can extract more details, such as decorations on the wall, while the CRBM lost most details of chair, carpet, and so on. Obviously, the CCRBM is better than the CRBM in feature extraction.

In the second experiment, in order to evaluate the qualities of the visualization of weights, we sampled 70x70 pixels area as input, and got hidden layer output with 60x60 pixels by averaging 24 groups. The trained weight matrices of CRBM and CCRBM are visualized in Fig.8. The left group is the results of CRBM and the right one is the results of

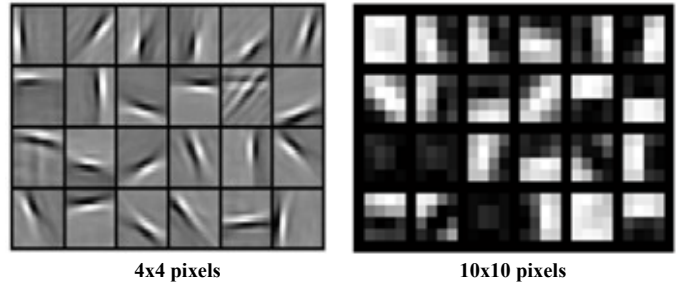
**4x4 pixels**   **10x10 pixels**

**Fig.10**.Left is the group with the size of 4x4 pixels. Right is the group with the size of 16x16 pixels. In the left image, the features are more distinct and smooth. In the right image, almost all features are fuzzy; even they can't be regarded as edges. Although the smaller size of filter is better in feature extraction, it cost much more time in computational.

CCRBM. The visualization of the weight matrix indicates the visible units would be maximally activated by the hidden units. As shown in Fig.8, the CCRBM can extract distinct features without much noise at 50 iterations. In Fig.9, we displayed the reconstruction error, i.e., the absolute value of the difference between the input data and the reconstruction data, which indicates the generative ability of the model and the accuracy of the reconstruction. It is evident from Fig.9 that the CCRBM performs better than the standard CRBM in the convergence speed and the final stability.

In the experiments, the size of each group affects the quality of extracted features and accuracy of classification. Small size of filter yields more details and hidden information in the extracted features, however, small size filter will increase the computation burden. Thus, it is important to choose an appropriate size of filters. In this paper, we conducted our experiments with different sizes of filters for the CCRBM; and the trained weight matrix are shown in Fig.10, from which we can see that the extracted features with 24 groups of 4x4 filters are very distinct, but the computation time is tripled than that using 10x10 filters.

The third experiment is to evaluate the performance of the CCDBN in scene recognition. We constructed a four layer CCDBN to recognize the object images in the Caltech 101 dataset. We randomly chose 15 training images per class and 5 testing images per class from Caltech-101dataset. First, we trained a CCRBM with 24 groups of 10x10 convolution filter and a 2x2 pooling filter. Then, we added the same structure for the second layer. After 500 iterations, the recognition rates of the first layer and the second layer are 56.4% and 61.8% respectively, as shown in Table 1, the results are better than those of the standard CRBM with the same architecture [18], Ranzato's method [27] and SVM-KNN model [28]. Finally, we trained the third layer and fourth layer with 24 groups of 6x6 convolution filter and a 2x2 pooling filter. We also got the better results than [29]. The results are shown in Table.1. The proposed CCDBN with four layers performs slightly worse than [30], however, the result of the counterpart was received by combing feature extraction algorithm with several post-processing procedures. [30] took advantage of non-negative sparse coding and low-rank and sparse matrix after feature extraction and utilized locality-constrained linear coding updating image representation.

*Table.1.* Classification accuracy over Caltech-101dataset

| Training Size | Recognition Rate |
|---|---|
| CCDBN(1 layer) | 56.4% |
| CCDBN(2 layers) | 62.8% |
| CCDBN(3 layers) | 67.7% |
| **CCDBN(4 layers)** | **73.6%** |
| CDBN(1 layer) | 53.2% |
| CDBN(2 layers) | 57.7% |
| Ranzato et al.[27] | 54% |
| Zhang et aL.[28] | 59% |
| Wang et al. [29] | 73.4% |
| Zhang et al. [30] | 75.6% |

*Table.2.*The Recognition Rate over MIT-indoor Scene Dataset

| Training Size | Recognition Rate |
|---|---|
| CCRBM | 15.7% |
| **CCDBN(4 layers)** | **41.2%** |
| CRBM | 13.2% |
| CDBN(4 layers) | 36.4% |
| ROI | 26.1% |
| DPM | 30.4% |
| Hybird-Parts | 39.8% |
| LPR | 44.8% |

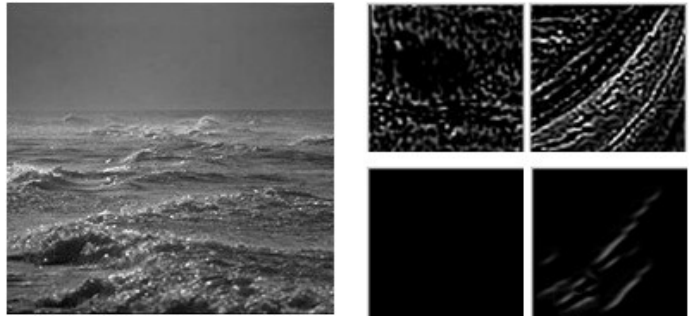

**Fig.11.** Left: an input coast image. Right: the image batches as the input data (top) and the reconstruction images by the features extracted by the centered CRBM (bottom).

In addition, we chose 50 training images per class and 20 testing images per class from MIT-indoor scene dataset and trained another CCDBN with four layers. Every layer consisted of 40 groups of 10x10 pixels filters and the pooling region of 2x2 pixels. The trained CCRBM with a softmax classifier was used to recognize scenes on the MIT-indoor scene dataset. The recognition rates are shown in Table.2. CCRBM [31] performs better than the standard CRBM. And the proposed CCDBN with four layers performs better than most counterparts, such as ROI [2], DPM [32], and Hybrid-Parts. In Table.2 the recognition rate of four-layer CCDBN was slightly lower than that of LPR [33], but our method is more applicable to practical issue, such as robot location and navigation since CDDBN is an unsupervised method and it can be applied to extract features from any image without label. Once the parameters are successfully estimated, the CCDBN model can be used for all kinds of images. While the parameters of LPR are unique for each kind of scene; and parameters must be trained for another type of scene every times. Thus, in practice, the proposed CDDBN is widely applicable than the LPR.

**5.2 Discussion**

Scene recognition is a challenging task in computer vision, while feature extraction is a crucial step of the recognition procedure. In recent years, deep learning models such as DBN, DBM, and CNN have attracted more and more attention. As an efficient generative model for full-sized natural images, the CRBM has been successfully used in handwritten recognition, object classification, and pedestrian detection. In this paper, we presented a CCRBM algorithm to cope with the instability of the CRBM caused by approximation and replacement. The CCRBM improves the performance by introducing centering factors during the learning procedure. The experimental results based on the fifteen scene categories dataset indicated that the CCRBM can obtain more distinct and detailed features than the standard CRBM. In addition, the proposed approach is computational efficient thanks to the centering factors. The experiments on Caltech 101 dataset also demonstrated that our method performs better than the standard CRBM. Moreover, to evaluate the performance of CCDBN stacked by CCRBMs in scene recognition, we constructed a four-layer CCDBN to recognize scene images. The experimental results on Caltech 101 dataset showed that our method performs better than other counterparts. In our experiments, however, we can find that there are some near-null images in the reconstruction results, as shown in the lower middle column of Fig.11. This is because the original image patches, such as coast, mountain and forest image patches, have few changes of grey values, thus, the values of extracted features will become zero or one after binarization.

## 6. Conclusions

In this paper, we have presented a CCRBM model by introducing the centering factors into the learning process of the CRBM. Compared to the standard CRBM, the CCRBM effectively reduces the sources of the instability caused by approximation and the structure of the model. The CCRBM can be widely applied to natural scene images, even for large size images, which is crucial in practical applications. Experimental results over fifteen scene categories, MIT-indoor scene dataset, and Caltech 101 datasets show that the CCRBM can obtain more distinct and detailed features than the standard CRBM. The experiments over MIT-indoor scene dataset and Caltech 101 dataset demonstrate that the CCDBN performs better than other counterparts in terms of accuracy of scene recognition. Although the proposed method has achieved promising results, further research is needed to find a more efficient approach to feature extraction and scene recognition.

**Acknowledgements**

This work is partly supported by the National Natural Science Foundation of China under Grant nos. 61201362, 61273282 and 81471770, the Scientific Research Project of Beijing Educational Committee under Grant no.KM201410005005.